\documentclass{article}
\usepackage{geometry}
\usepackage{amsthm, amssymb, bbm, mathrsfs}
\usepackage{statmath}
\usepackage{authblk}
\usepackage[utf8]{inputenc}
\usepackage{esint}
\usepackage{lipsum}
\usepackage{natbib}

\usepackage[english]{babel}
\theoremstyle{definition}
\newtheorem{definition}{Definition}[section]
\newtheorem{remark}[definition]{Remark}
\newtheorem{assumption}[definition]{Assumption}

\theoremstyle{plain}
\newtheorem{theorem}[definition]{Theorem}
\newtheorem{lemma}[definition]{Lemma}

\theoremstyle{plain}
\newtheorem{corollary}[definition]{Corollary}

\usepackage{amsmath}
\usepackage{graphicx}

\usepackage[colorlinks=true, allcolors=blue]{hyperref}

\title{{Limit Theorems for Stochastic Gradient Descent in High-Dimensional Single-Layer Networks}}
\author{Parsa Rangriz\thanks{prangriz@ucsd.edu\\This work was done while the author was affiliated with the Department of Statistics and Actuarial Science, University of Waterloo, Canada.}}
\affil{Department of Mathematics, University of California San Diego, La Jolla, CA 92093}

\date{}
\begin{document}
\maketitle
\begin{abstract}
    This paper studies the high-dimensional scaling limits of online stochastic gradient descent (SGD). Building on the work of Ben Arous, Gheissari, and Jagannath on the effective dynamics of SGD, we study the critical scaling regime of the step size for single-layer networks. Below this regime the effective dynamics are governed by deterministic (ballistic) limits, whereas at the critical scale a correction term emerges that changes the phase diagram. Near the fixed points of these dynamics, we show that the diffusive (SDE) limit of the rescaled correlation is an Ornstein--Uhlenbeck process. More precisely, it is mean-reverting whenever the information exponent is at least three. At information exponent two the drift has no universal sign, and the fixed point may become repelling; we show this explicitly for phase retrieval, where the sign is determined by the step size and the noise level. These results illustrate the limitations of deterministic scaling limits in capturing stochastic fluctuations in high-dimensional learning dynamics.
\end{abstract}
\section{Introduction}
Stochastic gradient descent (SGD) is one of the most widely used algorithms in machine learning, optimization, and data science. Since its introduction by Robbins and Monro in the 1950s \cite{robbins_stochastic_1951}, a main challenge in the theory of machine learning has been to understand how SGD navigates the non-convex loss landscape to train neural networks. While early works focused on fixed-dimensional settings, e.g., \cite{mcleish_functional_1976, benaim_dynamics_1999, benveniste_adaptive_1990, dupuis_stochastic_1989, kushner_asymptotic_1983, ljung_analysis_1977}, recent work has shifted toward high-dimensional regimes, where both the sample size and parameter dimension grow, e.g., \cite{needell_stochastic_2014, li_diffusion_2017, mandt_stochastic_2017, harvey_tight_2019, tan_online_2019}.

In fixed dimensions, the behavior of SGD can be studied using classical asymptotic methods and stochastic approximation theory, introduced by McLeish (1976) \cite{mcleish_functional_1976}, including pathwise limit theorems such as the functional central limit theorem (FCLT) and large deviation principles. In the regime of sufficiently small step size (learning rate) of the algorithm, with a fixed loss function, the scaling limit of the SGD trajectory has been shown to converge to the solution of a gradient flow problem, e.g., \cite{araujo_mean-field_2019, chizat_global_2018, mei_mean_2018, rotskoff_trainability_2022, sirignano_mean_2020}. There has been also growing interest in higher-order works via diffusion approximations, including asymptotic expansions of the SGD trajectory in terms of the step size, e.g., \cite{anastasiou_normal_2019, li_diffusion_2017, li_stochastic_2019, li_validity_2021}.

By contrast, in high-dimensional settings, tracking the full SGD trajectory is often infeasible. A common approach is to analyze scaling limits of lower-dimensional summary statistics under regularity and simplifying assumptions. A major development in this direction came in the late 1990s, when Saad and Solla \cite{saad_exact_1995, saad_-line_1999} introduced order parameters, such as the overlap matrix in the single-index model, drawing inspiration from the statistical physics of spin glasses. This perspective, known as dynamical mean-field theory (DMFT), has since provided a powerful framework for analyzing the high-dimensional dynamics of SGD, e.g., \cite{biehl_learning_1995, goldt_dynamics_2019, celentano_high-dimensional_2026, veiga_phase_2022, collins-woodfin_hitting_2024}.

Building on this breakthrough, subsequent research has focused on characterizing the classes of functions that SGD can efficiently learn, in terms of both time and sample complexity. The dynamics are typically studied in the ballistic phase, where summary statistics evolve on a macroscopic scale and are well-approximated by an ordinary differential equation (ODE). In this regime, the macroscopic behavior follows a deterministic scaling limit akin to the population gradient flow, e.g., \cite{goldt_dynamics_2019, veiga_phase_2022, collins-woodfin_hitting_2024}. 

In single-index models, for example, the time required for learning scales with the dimension, which depends sensitively on the geometry of the loss landscape. To analyze the dynamics near fixed points of these ODEs, Ben Arous, Gheissari, and Jagannath \cite{arous_online_2021} introduced the concept of the information exponent, a geometric quantity that captures how SGD explores the loss landscape. They showed that there are three distinct behaviors in which the time to weakly recover is linear, quasi-linear, or polynomial, depending on whether the information exponent is less than two, equal to two, or greater than two, respectively.

In this paper, we establish a FCLT for the rescaled dynamics of SGD in single-index models. Specifically, we analyze the diffusive phase, where the summary statistics fluctuate microscopically around fixed points, and the ballistic approximation breaks down. In a critical scaling regime for the step size, an additional correction term arises in the dynamics, leading to significant deviations from the population gradient flow. In microscopic neighborhoods of a fixed point, the effective dynamics become stochastic and are governed by stochastic differential equations (SDEs), which may display a wide range of behaviors, including degenerate cases. 

Our main focus is on learning single-index models whose activation has information exponent at least two. We show that, under random initialization, the high-dimensional trajectory of SGD with stochastic corrections deviates from the deterministic limit with no population corrector predicted by DMFT. At the critical step size the correlation stays at zero along the ballistic limit, so the recovery behavior is visible only in its rescaled fluctuations. The dynamics converge to an Ornstein–Uhlenbeck (OU) process, and the information exponent decides the sign of its drift: mean-reverting when the exponent is at least three, of either sign when it equals two. We illustrate this for phase retrieval.

\section{Setting and Assumptions}\label{main}
Suppose that we are given a parametric family of distributions, $(\mathbb P_x)_{x \in \mathbb R^{N}}$. According to the teacher-student scenario, the teacher begins by generating a hidden vector $x^* \in \mathbb{R}^N$ from a known prior distribution. Based on a statistical model, the teacher then produces a sequence of i.i.d. observations $(Y_k)_k$, each generated conditionally on $x^*$ and parameterized by elements in $\mathcal{Y}_N \subseteq \mathbb{R}^N$, from $P_N = \mathbb P_{x^*}$, which we call the \textit{data distribution}. The number of observations is indexed by $k \in \{1, \dots, N\}$. The teacher then provides the dataset $(Y_k)_k$, along with partial information about the generative model, to the student. The student’s objective is to infer the hidden variables $x^*$ using only the observed data and the provided model information.

Suppose a sequence of parameter iterates $(X_k)_k$ lies in a high-dimensional space $\mathcal{X}_N \subseteq \mathbb{R}^N$, and the training data $(Y_k)_k$ takes values in $\mathcal{Y}_N \subseteq \mathbb{R}^N$. The learning proceeds via online SGD with respect to a loss function $L_N : \mathcal{X}_N \times \mathcal{Y}_N \to \mathbb{R}$, and a constant step-size $\delta_N = c_\delta/N$ (for a fixed constant $c_\delta > 0$ independent of $N$), according to the update rule
\begin{equation*}
    X_{k+1} = X_k - \delta_N \nabla L_N(X_k; Y_{k+1}),
\end{equation*}
initialized with a random vector $X_0 \sim \mu_N \in \mathcal{M}_1(\mathbb{R}^N)$, where $\mathcal{M}_1(\mathbb{R})$ denotes the space of probability measures on $\mathbb{R}$. Our goal is to understand the evolution of the sequence $(X_k)$ in the high-dimensional limit as $N \to \infty$. To this end, suppose that we are given a sequence of functions $\mathbf u_N \in C^1(\mathbb R^N; \mathbb R^l)$ for some fixed $l$, where $\mathbf u_N(x) = (u_1^N(x), \dots, u_l^N(x))$, and more precisely, our goal is to understand the evolution of $\mathbf u_N(X)$. 

To develop a scaling limit, we need some regularity assumptions on the relationship between how the step-size $\delta_N$ scales in relation to the loss $L_N$, its gradients, and the data distribution $P_N$. Therefore, we define the \textit{sample-wise error} as follows
\begin{equation*}
    H(x, Y) := L_N(x, Y) - \Phi(x) \quad \text{where} \quad \Phi(x) := \mathbb E[L_N(x, Y)].
\end{equation*}

In the following, we suppress the dependence of $H$ on $Y$ and instead view $H$ as a random function of $x$, which we denote as $H(x)$. We let $V(x) = \mathbb E[\nabla H(x) \otimes \nabla H(x)]$ be the covariance matrix for $\nabla H$ at $x$.

Consider the following model of supervised learning with a \textit{single-layer network}\footnote{This model and special cases thereof have been studied under many different names by a broad range of communities: single-layer neural networks, teacher-student networks, and single-index models.}: Suppose we are given a (possibly) non-linear activation function $f: \mathbb R \to \mathbb R$, a set of feature vectors $(a_k)_k$, and additive noisy responses $(\epsilon_k)_k$ of the form
\begin{equation*}
    y_k = f(\langle a_k, x^*\rangle) + \epsilon_k,
\end{equation*}
and for the sake of simplicity, we consider quadratic loss functions
\begin{equation*}
    L_N(x, Y) = L_N(x, (y, a)) = (y - f(\langle a, x))^2.
\end{equation*}

Let us focus on the most studied regime, namely where $(a_k)_k$ are i.i.d. standard Gaussian vectors in $\mathbb R^N$; for the $(\epsilon_k)_k$ we only assume they are i.i.d. mean zero with variance $C_\epsilon$ and finite $4 + \delta$-th moment for some $\delta > 0$. 

Note that we may write the population loss as
\begin{equation*}
    \Phi(x) = \mathbb E\left[\left(f(\langle a, x\rangle) - f(\langle a, x^*\rangle)\right)^2\right] + C_\epsilon.
\end{equation*}
Also, in our case, the sufficient number of summary statistics we need for the single-index model is two, and let $x^* \in \mathbb S^{N-1}$ be a fixed unit vector. Now, we define the following summary statistics $\mathbf u_N(x) = (u_1^N(x), u_2^N(x))$
\begin{equation*}
    u_1^N(x) := m(x) = \langle x, x^*\rangle, \qquad u_2^N(x) := r_\perp^2(x) = \|x\|^2 - m^2(x),
\end{equation*}
where the loss distribution only depends on $(m, r_\perp^2)$ and the population loss is of the form $\Phi(x) =  \phi(m,r_\perp^2)$. We call $m$ the \textit{correlation} of $x$ with $x^*$. We also refer to $r_\perp > 0$ as the \textit{radius}.
\begin{remark}\label{invariance}
    Note that the model is unchanged by the substitution $f \mapsto f + c$ for any constant $c \in \mathbb R$. If the teacher's activation is shifted, the observed responses become $y_k = (f+c)(\langle a_k, x^*\rangle) + \epsilon_k$, and the student's loss $L_N(x, Y)$ under the shift is invariant. Consequently, the loss $L_N$, the population loss $\Phi$, the sample-wise error $H$, and the covariance $V$ are invariant under the shift. 
\end{remark}
To ensure the tightness of the trajectories of the summary statistics, \cite{arous_high-dimensional_2024} impose two key assumptions, namely \textit{localizablity} and \textit{asymptotic closability} on the triplet $(\mathbf u_N, L_N, P_N)$ and the learning rate $\delta_N$.
\begin{definition}\label{loc_def}
   A triple $(\mathbf u_N, L_N, P_N)$ is $\delta_N$-localizable with localizing sequence $(E_K)_K$ if there is an exhaustion by compacts $(E_K)_K$ and constant $C_K$ (independent of $N$) such that
   \begin{enumerate}
       \item $\max_i \sup_{x \in \mathbf u_N^{-1}(E_K)} \|\nabla^2 u_i^N\|_{\text{op}} \leq C_K \delta_N^{-1/2}$, and $\max_i \sup_{x \in \mathbf u_N^{-1}(E_K)} \|\nabla^3 u_i^N\|_{\text{op}} \leq C_K$.
       \item $\sup_{x \in \mathbf u_N^{-1}(E_K)} \|\nabla \Phi\| \leq C_K$ and $\sup_{x \in \mathbf u_N^{-1}(E_K)} \mathbb E[\|\nabla H\|^8] \leq C_K \delta_N^{-4}$.
       \item $\max_i \sup_{x \in \mathbf u_N^{-1}(E_K)} \mathbb E[\langle \nabla H, \nabla u_i^N\rangle^4] \leq C_K \delta_N^{-2},$ and \\$\max_i \sup_{x \in \mathbf u_N^{-1}(E_K)}  \mathbb E[\langle \nabla^2 u_i^N, \nabla H \otimes \nabla H - V\rangle^2] = o(\delta_N^{-3})$
   \end{enumerate}
\end{definition}
\begin{definition}
    A family of summary statistics $(\mathbf u_N)$ are asymptotically closable for learning rate $\delta_N$ if $(\mathbf u_N, L_N, H)$ are $\delta_N$-localizable with localizing sequence $(E_K)_K$, and furthermore there exist locally Lipschitz functions $\mathcal H: \mathbb R^l \to \mathbb R^l$ and $\Sigma: \mathbb R^l \to \mathbb R^{l \times l}$, such that
    \begin{equation*}\label{asympt}
        \sup_{x \in \mathbf u_N^{-1}(E_k)}  \|(-\mathcal A_N + \delta_N \mathcal L_N) \mathbf u_N(x) - \mathcal H(\mathbf u_N(x))\| \to 0,
    \end{equation*}
    and
    \begin{equation*}\label{asymptII}
        \sup_{x \in \mathbf u_N^{-1}(E_k)}  \|\delta_N J_N V J_N^T - \Sigma(\mathbf u_N(x))\| \to 0
    \end{equation*}
where $\mathcal A_N = \langle \nabla \Phi, \nabla\rangle$ and $\mathcal L_N = \frac{1}{2}\langle V, \nabla^2 \rangle$. We call $\mathcal H$ the \textit{effective drift}, and $\Sigma$ the \textit{effective volatility}.
\end{definition}

For the sake of simplicity, suppose that not only the asymptotic closability is satisfied, but each of the two terms $\mathcal A_N  \mathbf u_N$ and $\delta_N \mathcal L_N  \mathbf u_N$ in Definition \ref{asympt} individually admit $N \to \infty$ limits: namely there exist $\mathcal F, \mathcal G: \mathbb R^l \to \mathbb R^l$ such that 
\begin{equation*}
    \sup_{x \in  \mathbf u_N^{-1}(E_K)}\|\mathcal A_N  \mathbf u_N(x) - \mathcal F( \mathbf u_N(x))\| \to 0, 
\end{equation*}
and
\begin{equation*}
    \sup_{x \in  \mathbf u_N^{-1}(E_K)}\|\delta_N \mathcal L_N  \mathbf u_N(x) - \mathcal G( \mathbf u_N(x))\| \to 0,
\end{equation*}
evidently $\mathcal H = -\mathcal F + \mathcal G$, and we call $\mathcal F \text{ and } \mathcal G$ the \textit{population drift} and the \textit{population corrector}, respectively.

Then the corresponding (possibly stochastic) differential equation of online SGD \cite{arous_high-dimensional_2024} is given by
    \begin{equation}\label{aukosh}
        \mathrm d\mathbf u_t = (-\mathcal F(\mathbf u_t) + \mathcal{G}(\mathbf u_t)) \mathrm dt + \sqrt{\Sigma( \mathbf u_t)} \mathrm d \mathbf B_t,
    \end{equation}
where $\mathbf B_t$ is a standard Brownian motion in $\mathbb R^l$.

Recall that the {Hermite polynomials}, which we denote by $(h_k(x))_{k = 0}^\infty$, are the normalized orthogonal polynomials of the Gaussian distribution $\gamma(x) \propto \exp(-x^2/2) dx$. Define the $k$-th Hermite coefficient for an activation function $f \in L^2(\gamma)$ by
\begin{equation*}
    \alpha_k := \langle f, h_k\rangle_{L^2} = \frac{1}{\sqrt{2\pi}} \int_{-\infty}^\infty f(z) h_k(z) e^{-z^2/2} dz.
\end{equation*}
Also, we define the norm $\|f\|_{L^2}^2 := \langle f, f\rangle_{L^2}$. As long as $f'$ has at most polynomial growth, the population loss is differentiable, and \ref{aukosh} exists. 

According to \cite{arous_online_2021}, under the regularity conditions previously discussed, in the following we define a key quantity governing the performance of online SGD. 
\begin{definition}
    We say that an activation function $f: \mathbb R \to \mathbb R$ has information exponent $k$ if the first non-zero coefficient in its Hermite expansion is the $k$th coefficient, i.e., $\alpha_i = 0$ for all $1 \leq i < k$ and $\alpha_k \neq 0$.
\end{definition}

\section{Main Results}
In this work, we focus exclusively on activation functions with an information exponent of at least two. This choice implies that the corresponding sample complexity of online SGD grows at least quasi-linearly—or potentially polynomially—with the dimension, placing our analysis squarely in the more challenging high-dimensional learning regime.

Even with this relative simplicity, we encounter various ODE and SDE limits following the general form of (\ref{aukosh}). Indeed, we find dynamical phase transitions corresponding to the aforementioned threshold in our model. Our analysis focuses exclusively on the most interesting, critical step-size scaling $\delta_N = c_\delta/N$ corresponding to the proportional asymptotics regime from the random matrix theory literature. 

Before stating our main results, we introduce a notation for the Gaussian functionals of $f$ that recur throughout, and the mentioned regularity assumptions on $f$. For $r_\perp > 0$ and $a \sim \mathcal N(0,1)$, write
\begin{align}\label{shorthand}
    \nonumber \mathsf a(r_\perp) &:= \mathbb E[f'^2(ar_\perp)], \qquad
    \mathsf b(r_\perp) := \mathbb E[f(ar_\perp)f''(ar_\perp)], \qquad
    \mathsf s(r_\perp) := \mathbb E[f''(ar_\perp)],\\
    \mathsf t(r_\perp) &:= \mathbb E[f'^2(ar_\perp)f(ar_\perp)], \qquad
    \mathsf q(r_\perp) := \mathbb E[f'^2(ar_\perp)f^2(ar_\perp)].
\end{align}
\begin{assumption}\label{assumptions}
    Assume that the activation function $f$ holds the following conditions:
    \begin{enumerate}
        \item[(A1)] $f \in L^2(\gamma)$ is twice-differentiable a.e. and $f', f''$ have at most polynomial growth;
        \item[(A2)] the information exponent of $f$ is at least two.
    \end{enumerate}
\end{assumption}

\subsection{Sub-critical scaling regime}
We are now ready to present our main results. 

\begin{theorem}\label{ode}
 Let $(X_k^{\delta_N})_k$ be SGD initialized from $X_0 \sim \mu_N$ for $\mu_N \in \mathcal M_1(\mathbb R^{N})$ with the learning rate $\delta_N$ for the quadratic loss $L_N$ of a single-index model. Suppose that the activation function $f$ satisfies Assumption \ref{assumptions}. For the corresponding summary statistics (correlation and radius) $ \mathbf u_N = (u_i^N)_{i = 1}^2 = (m, r_\perp^2)$, let $( \mathbf u_N(t))_t$ be the linear interpolation of $( \mathbf u_N(X_{\lfloor t \delta_N^{-1} \rfloor}^{\delta_N}))_t$. Then $\mathbf u_N$ are asymptotically closable with learning rate $\delta_N = c_\delta/N$. Moreover, $ \mathbf u_N = (m, r_\perp^2)$ converges as $N \to \infty$ to the solution of the following ODE initialized from the pushforward of the initial data $\lim_{N \to \infty} ( \mathbf u_N)_* \mu_N$,
    \begin{align}
\begin{cases}
\displaystyle \frac{\mathrm{d}m}{\mathrm{d}t} = -2\mathbb E_{a_1, a_2}[a_1 f'(a_1 m + a_2 r_\perp)(f(a_1 m + a_2 r_\perp) - f(a_1))] ,\\[0.5em]
\displaystyle \frac{\mathrm dr_\perp^2}{\mathrm dt} = -4\mathbb E_{a_1, a_2}[a_2r_\perp f'(a_1 m + a_2 r_\perp)(f(a_1 m + a_2 r_\perp) - f(a_1))] \\
\qquad \qquad + 4c_\delta\mathbb E_{a_1, a_2}[f'^2(a_1 m+ a_2 r_\perp)((f(a_1 m+ a_2 r_\perp) - f(a_1))^2 + C_\epsilon)],
\end{cases} \label{effective_m}
\end{align}
    where $a_1, a_2$ are i.i.d. standard Gaussian variables.
\end{theorem}
We can obtain the following result when we restrict the initialization of the algorithm to be chosen randomly from a gaussian distribution, i.e., $X_0 \sim \mathcal N(0, \frac{\sigma^2}{N}I_N)$, then $(\mathbf u_N)_* \mu_N \to \delta_{(0, \sigma^2)}$ weakly for some fixed $\sigma^2$.
\begin{corollary}\label{cor1}
     Under the conditions of Theorem \ref{ode}, $r_\perp^2$ converges as $N \to \infty$ to the solution of the following ODE initialized from the pushforward of the initial data $\lim_{N \to \infty} ( \mathbf u_N)_* \mu_N = \delta_{(0, \sigma^2)}$,
    \begin{equation}\label{confine_heat}
        \frac{\mathrm d r_\perp^2}{\mathrm dt} = -4r_\perp^2\,\gamma(r_\perp) + 4c_\delta\,\mathsf n(r_\perp),
    \end{equation}
     where
    \begin{equation*}
    \begin{cases}
        \gamma(r_\perp) = \mathsf a(r_\perp) + \mathsf b(r_\perp) - \alpha_0 \mathsf s(r_\perp) =  \mathbb E_{a_2}[f'^2(a_2 r_\perp) + (f(a_2 r_\perp)-\alpha_0)f''(a_2 r_\perp)]\\
        \mathsf n(r_\perp) := \mathbb E_{a_2}\big[f'^2(a_2r_\perp)\big(f(a_2r_\perp)-\alpha_0\big)^2\big]
        + \mathsf a(r_\perp)\big(C_\epsilon + \|f-\alpha_0\|_{L^2}^2\big) \;>\; 0 .
    \end{cases}
    \end{equation*}
    Moreover, $m \equiv 0$ at all time, where the functionals $\mathsf a,\mathsf b,\mathsf s$ are as in \eqref{shorthand}.
\end{corollary}
\begin{remark}\label{existence}
    Whether the ODE (\ref{confine_heat}) admits a fixed point depends jointly on the activation
    $f$ and on the step-size constant $c_\delta$. 
    The first term of (\ref{confine_heat}) is the confinement produced by the population drift
    $\mathcal F_{r_\perp^2}$, while the second is the heating produced by the corrector $\mathcal G_{r_\perp^2}$; only the
    latter carries $c_\delta$. Since $\mathsf n(r_\perp) > 0$, a radius $r_\perp^* > 0$ is a fixed point if and
    only if $c_\delta = \Psi(r_\perp^*)$, where
    \begin{equation*}
        \Psi(r_\perp) := \frac{r_\perp^2\,\gamma(r_\perp)}{\mathsf n(r_\perp)} .
    \end{equation*}
    
    Hence a fixed point exists precisely when $c_\delta$ lies in the range of $\Psi$, and in
    particular only when $c_\delta \leq c_\delta^{\mathrm{crit}} := \sup_{r_\perp>0}\Psi(r_\perp)$; above this
    threshold the heating dominates at every radius and $r_\perp^2$ diverges. 
    \end{remark}
\subsection{Critical scaling regime}
Let us consider a rescaling regime of $\mathbf u_N$ in a microscopic neighborhood of the fixed point $m = 0$. This captures the initial phase from a random start if $\mu_N\sim \mathcal N(0, \frac{\sigma^2}{N} I_N)$ for some fixed $\sigma^2 > 0$. Then the pushforward satisfies then $(\mathbf u_N)_*\mu_N \to \mathcal \delta_{(0, \sigma^2)}$ weakly. Now rescale and let $\tilde{\mathbf{u}}_N = (\tilde m, r_\perp^2)$ where $\tilde m = \sqrt N m$. Evidently, $\bar \nu = \lim_{N \to \infty} (\tilde{\mathbf{u}}_N)_* \mu_N = \mathcal N(0, \sigma^2) \otimes \delta_{\sigma^2}$.
\begin{theorem}\label{sde}
     Let $(X_k^{\delta_N})_k$ be SGD initialized from $X_0 \sim \mathcal N(0, \frac{\sigma^2}{N}I_N)$ for some fixed $\sigma^2$ with learning rate $\delta_N$ for the quadratic loss $L_N$ of a single-index model. Suppose that the activation function $f$ satisfies Assumption \ref{assumptions}. For the corresponding summary statistics (rescaled correlation and radius) $ \tilde{\mathbf{u}}_N = (\tilde u_i^N)_{i = 1}^2 = (\tilde m, r_\perp^2)$, let $(\tilde{\mathbf{u}}_N(t))_t$ be the linear interpolation of $(\tilde{\mathbf{u}}_N(X_{\lfloor t \delta_N^{-1} \rfloor}^{\delta_N}))_t$. Then $\tilde{\mathbf{u}}_N$ are asymptotically closable with learning rate $\delta_N = c_\delta/N$. Moreover, $\tilde{\mathbf{u}}_N = (\tilde m, r_\perp^2)$ converges as $N \to \infty$ to the solution of the following SDE initialized from $\mathcal N(0, \sigma^2) \otimes \delta_{\sigma^2}$,
\begin{equation} \label{sde_r}
\begin{cases}
\mathrm{d}\tilde{m} = -2\tilde{m}\,\big(\gamma(r_\perp) - \alpha_2 \mathsf s(r_\perp)\big) \mathrm{d}t + 2\sqrt{c_\delta\big(\mathsf{q}(r_\perp) - 2(\alpha_0+\alpha_2)\mathsf{t}(r_\perp) + \mathsf{a}(r_\perp)\eta\big)}\,\mathrm{d}B_t,\\[0.5em]
\dfrac{\mathrm{d} r_\perp^2}{\mathrm{d}t} = -4r_\perp^2\,\gamma(r_\perp) + 4c_\delta\,\mathsf{n}(r_\perp),
\end{cases}
\end{equation}
     where
    \begin{equation*}
        \begin{cases}
            \gamma(r_\perp) = \mathsf a(r_\perp) + \mathsf b(r_\perp) - \alpha_0 \mathsf s(r_\perp) =  \mathbb E_{a_2}[f'^2(a_2 r_\perp) + (f(a_2 r_\perp)-\alpha_0)f''(a_2 r_\perp)]\\
            \eta = \|f\|_{L^2}^2 + 2\|f'\|_{L^2}^2 + 2\langle f, f''\rangle_{L^2} + C_\epsilon,\\
            \mathsf n(r_\perp) := \mathbb E_{a_2}\big[f'^2(a_2r_\perp)\big(f(a_2r_\perp)-\alpha_0\big)^2\big]
        + \mathsf a(r_\perp)\big(C_\epsilon + \|f-\alpha_0\|_{L^2}^2\big)
        \end{cases}
    \end{equation*}
    with $a_1, a_2$ i.i.d.\ standard Gaussian, $\alpha_0 = \mathbb E_{a_1}[f(a_1)]$, and $\alpha_2 = \mathbb E_{a_1}[f''(a_1)]$ and the functionals $\mathsf a,\mathsf b,\mathsf s,\mathsf t,\mathsf q$ are as in \eqref{shorthand}.
\end{theorem}
    By Remark \ref{invariance}, since the SGD trajectory $(X_k)_k$ is driven through $\nabla L_N$, every limiting ODE and SDE derived above is therefore obligated to be invariant as well. 

Now, if the information exponent is at least three and the radius term of the rescaled summary statistics starts from the initial point $r_\perp^*$, then a \textit{mean-reverting} OU process appears.
\begin{corollary}\label{cor2}
    Assume the conditions of Theorem \ref{sde} hold, except that the information exponent of $f$ is now at least three, i.e., $\alpha_2 = 0$. If $\mu_N \sim \mathcal N(0, \frac{{r_\perp^*}^2}{N} I_N)$ where ${r_\perp^*}^2$ is the fixed-point of the ODE for $r_\perp^2$ of (\ref{sde_r}), then $\tilde m$ converges as $N \to \infty$ to the solution of the following mean-reverting OU process, i.e., $\gamma(r_\perp^*) > 0$,
    \begin{align}\label{cor2_sde}
        \mathrm{d}\tilde{m} = -2\tilde{m}\,\gamma(r_\perp^*) \mathrm{d}t + 2\sqrt{c_\delta\big(\mathsf{q}(r_\perp^*) - 2\alpha_0\mathsf{t}(r_\perp^*) + \mathsf{a}(r_\perp^*)\eta\big)}\,\mathrm{d}B_t.
    \end{align}
\end{corollary}
\begin{remark}\label{ie2}
    The mean-reverting OU process in Corollary \ref{cor2} is special to information exponent at least three, and does not necessarily hold at information exponent exactly two. If $\alpha_2 \neq 0$, the drift coefficient becomes $\theta(r_\perp^*) := \gamma(r_\perp^*) - \alpha_2 \mathsf s(r_\perp^*).$ Then, there is a competition between $\gamma(r_\perp^*)$ and $\alpha_2 \mathsf s(r_\perp^*)$, i.e., the fixed point $m = 0$ can be mean-reverting or mean-repelling depending on $f$, with no universal sign.
    \end{remark}
\subsection{Example: Phase Retrieval}
    Phase retrieval, $f(x) = x^2$, makes the dichotomy of Remark \ref{ie2} completely explicit, exhibiting both signs of the drift within a single model of information exponent two. By defining $u = r_\perp^2$, one gets
        $\Psi(u) = (3u-1)/\big(2(15u^2 - 6u + 3 + C_\epsilon)\big)$, whence
        \begin{equation*}
            c_\delta^{\mathrm{crit}} = \frac{\sqrt{5(8+3C_\epsilon)}-2}{4(12+5C_\epsilon)} ,
        \end{equation*}
        equal to $(\sqrt{10}-1)/24 \approx 0.0901$ when $C_\epsilon = 0$. Below the threshold the two
        roots of $\Psi(u) = c_\delta$ are a stable fixed point and an unstable barrier; as
        $c_\delta \downarrow 0$ the former decreases to the population value $u = 1/3$ and the latter
        escapes to infinity, while at $c_\delta = c_\delta^{\mathrm{crit}}$ they merge in a
        saddle-node bifurcation.

    Here
    $\alpha_0 = 1$, $\alpha_2 = 2$. Also,  $\mathsf a(r_\perp) = 4r_\perp^2$, $\mathsf b(r_\perp) = 2r_\perp^2$, $\mathsf s(r_\perp) = 2$, $\mathsf t(r_\perp) = 12 r_\perp^4$, $\mathsf q(r_\perp) = 60 r_\perp^6,$
    so that $\gamma(r_\perp) = 2(3r_\perp^2 - 1)$ while the drift of the rescaled correlation is
    governed by
    \begin{equation}\label{pr_theta}
        \theta(r_\perp) = \gamma(r_\perp) - \alpha_2 \mathsf s(r_\perp) = 6\big(r_\perp^2 - 1\big).
    \end{equation}
    
    Writing $u = r_\perp^2$, the RHS of (\ref{confine_heat}) becomes $4u\big(2 - 6u + 4c_\delta(15u^2 - 6u + 3 + C_\epsilon)\big)$, whose
    positive zeros are
    \begin{equation*}
        u_\pm = \frac{(24c_\delta + 6) \pm \sqrt{36 - 192 c_\delta - (2304 + 960 C_\epsilon)c_\delta^2}}{120\, c_\delta},
    \end{equation*}
    with $u_-$ stable and $u_+$ an unstable barrier. Initializing at $r_\perp^* = \sqrt{u_-}$, the
    rescaled correlation is then an OU process,
    \begin{equation}\label{pr_ou}
        \mathrm d \tilde m = -12\big({r_\perp^*}^2 - 1\big)\,\tilde m\, \mathrm dt
        + 4 r_\perp^*\sqrt{c_\delta\big(15 {r_\perp^*}^4 - 18 {r_\perp^*}^2 + 15 + C_\epsilon\big)}\,\mathrm dB_t,
    \end{equation}
    whose volatility never degenerates, as $15v^2 - 18v + 15 + C_\epsilon > 0$ for all $v \geq 0$.
    By (\ref{pr_theta}) the process is mean-reverting when ${r_\perp^*} > 1$, and evaluating
    the quadratic above at $u = 1$ shows that this happens if and only if
    \begin{equation*}
        C_\epsilon > 4 \qquad \text{and} \qquad
        \frac{1}{12 + C_\epsilon} < c_\delta \leq c_\delta^{\mathrm{crit}}
        = \frac{\sqrt{5(8+3C_\epsilon)}-2}{4(12+5C_\epsilon)} .
    \end{equation*}
    The two thresholds coincide at $C_\epsilon = 4$, where $c_\delta^{\mathrm{crit}} = 1/16$. Hence
    for $C_\epsilon \leq 4$ the correlation is mean-\emph{repelling} at the stable radius for every
    admissible step size, and in particular $\theta(r_\perp^*) \to -4$ as $c_\delta \downarrow 0$,
    since $u_- \downarrow 1/3$; only in a narrow window of large step sizes and large noise does the
    fixed point become mean-reverting. This is precisely the behaviour that Corollary \ref{cor2}
    rules out at information exponent at least three, and it is the single-index analogue of the
    signal-to-noise dependent drift found for matrix PCA in
    \cite{arous_high-dimensional_2024}.
\section{Discussion}
We showed that, with random initialization, the effective dynamics of SGD deviate from the trajectories predicted by deterministic limits in high-dimensional settings. As discussed earlier, DMFT describes a deterministic scaling limit via an ODE approximation akin to the population gradient flow. However, in the critical step-size regime, diffusive effects emerge, and the effective dynamics depart from this deterministic description due to the presence of a stochastic correction, which we refer to as the population corrector. At the critical step-size $\delta_N = c_\delta/N$, and when the information exponent is at least two, the correlation $m = \langle x, x^*\rangle$ stays at zero, which is a fixed point of the corresponding ballistic limit. This indicates that deterministic limits fail to fully capture the recovery behavior in this regime.

Near a fixed point of the radius equation, we showed that the rescaled correlation $\tilde m$ follows an OU process, and the information exponent decides whether this process pulls $\tilde m$ back to zero or pushes it away. If the information exponent is at least three, the drift always points back to zero, so $m = 0$ stays stable even once the stochastic correction is taken into account. If it is exactly two, the drift may point either way, depending on the activation function, and the fixed point can then be mean-repelling rather than mean-reverting.

As an example, we investigated phase retrieval, which shows how much the corrector can change the dynamics. Its population drift is by itself confining: this drift vanishes at $r_\perp^2 = 1/3$, so the gradient flow of $\Phi$ alone would leave the radius bounded. The heating comes entirely from the corrector, and once $c_\delta$ passes $c_\delta^{\mathrm{crit}}$ this control is lost and the radius grows without bound. The two signs of the drift also match what is known about sample complexity. At information exponent two, a drift that pushes $\tilde m$ away from zero lets the small initial correlation, of size $O(N^{-1/2})$, grow. This is why weak recovery needs only quasi-linearly many samples \cite{arous_online_2021}. At information exponent at least three, a drift that pulls $\tilde m$ back keeps the correlation trapped near the origin, so polynomially many samples are needed instead.

\bibliographystyle{plain}
\section{Proofs}\label{proofs}
\subsection{Proofs of Theorem \ref{ode} and Corollary \ref{cor1}}
\begin{lemma}\label{loc}
    In a single-index model, the distribution of the loss $L_N(x, (a, y))$ depends only on $ \mathbf u_N = (m, r_\perp^2)$. Also, $ \mathbf u_N$ is $\delta_N$-localizable for $E_K$ being the centered balls of radius $K$ in $\mathbb R^2$.
\end{lemma}
\begin{proof}
    We check the items in Definition \ref{loc_def}. By rotational invariance of the Gaussian ensemble, we may take $x^* = v$ where $v$ is the first basis vector of $\mathbb R^N$. First note that for every $x$, since $f$ is differentiable and $f'$ is of at most polynomial growth,
\begin{equation*}
\nabla \Phi = {\partial_m \phi} \nabla m +{\partial_{r^2_\perp} \phi} \nabla r_\perp^2,
\end{equation*}
where
\begin{equation*}
    \begin{cases}
        \partial_m \phi = 2\mathbb E_{a_1, a_2}[a_1 f'(a_1 m + a_2 r_\perp) (f(a_1 m + a_2 r_\perp) - f(a_1))],\\
    \partial_{r_\perp^2} \phi = \frac{1}{r_\perp}\mathbb E_{a_1, a_2}[a_2 f'(a_1 m + a_2 r_\perp) (f(a_1 m + a_2 r_\perp) - f(a_1))].
    \end{cases}
\end{equation*}
Note that, $(a_k)_{k=1}^N$ are i.i.d Gaussian variables as stated in the Introduction, but here by rotational invariance, we rename $a_2$ such that $\mathbb E[f(\langle a, x\rangle) ] = \mathbb E[f(a_1 m + a_2 r_\perp)]$. In particular, $a_2$ here does not correspond to the original second coordinate, but to the component orthogonal to $v$.

One may express the derivatives for $u_N$ as
\begin{equation*}
    \nabla m = v, \qquad \nabla r^2_\perp = 2(x - mv).
\end{equation*}
Notice that $\nabla^2 m = 0$, while $\nabla^2 r^2_\perp = 2(I - vv^T)$, and $\nabla^l m = \nabla^l r_\perp^2 = 0$ for all $l \geq 3$. It yields that 
\begin{equation*}
    \langle \nabla m, \nabla m\rangle = 1, \qquad \langle \nabla m, \nabla r^2_\perp\rangle = 0, \qquad \langle \nabla r^2_\perp, \nabla r_\perp^2\rangle = 4r_\perp^2.
\end{equation*}

For part (2), one may write
\begin{equation*}
    \|\nabla \Phi\| \leq |\partial_m \phi| \|\nabla m\| + |\partial_{r_\perp^2} \phi| \|\nabla r_\perp^2\|,
\end{equation*}
the bounding quantity is evidently a continuous function of $m, r^2_\perp$ and therefore as long as $x$ is such that $(m, r_\perp^2) \in E_K$, it is bounded by some constant $C_K$. 

Recall,
\begin{equation*}
    H(x, a) = (f(\langle a, x\rangle) - f(a_1))^2 -  \Phi(x).
\end{equation*}
Then the derivatives of $H$ are given by
\begin{equation*}
    \nabla H(x, a) = 2a f'(\langle a, x\rangle)(f(\langle a, x\rangle - f(a_1)) -\nabla \Phi(x).
\end{equation*}
Since $f'$ has at most polynomial growth, $\|a\| = O_p(\sqrt{N})$, and $\|\nabla \Phi\| = O_p(1)$, where $O_p$ is stochastic boundedness, we get
\begin{equation*}
    \|\nabla H\| \leq  2\|a\|\left|f'(\langle a, x\rangle) (f(\langle a, x\rangle) - f(a_1))\right| + \|\nabla \Phi\| = O_p(\sqrt{N}).
\end{equation*}
 Therefore, there exists $C_K(f) > 0$ independent of $N$ such that
\begin{equation*}
    \mathbb E[\|\nabla H\|^8] \leq C_K(f)N^4.
\end{equation*}
Moving on item (3), for every $w$,
\begin{equation*}
    \mathbb E[\langle \nabla H, w\rangle^4] \leq \mathbb E[\|\nabla H\|^4]  \|w\|^4\leq C_K(f) N^2.
\end{equation*}
If $w = \nabla m = v$, then $\|w\| = 1$ and if $w = \nabla r^2_\perp = 2(x - mv)$ then $\|w\| = 4r_\perp^2 \leq c_K$, for some constant $c_k$, so in both cases the upper bound is at most $C_K(f) N^2$. Furthermore,
\begin{align*}
    \mathbb E[\langle \nabla^2 r^2_\perp, \nabla H \otimes \nabla H - V\rangle^2] &\leq 4 \mathbb E[\langle(I - vv^T), \nabla H \otimes \nabla H - V\rangle^2 \leq  4 \mathbb E[\|\nabla H\|^4].
\end{align*}
The quantity $\mathbb E[\|\nabla H\|^4]$ is at most $N^2$ by the above proved second item in the definition of localizability. This is therefore $O_p(\delta_N^{-2}) = o(\delta_N^{-3})$ as claimed.
\end{proof}
\begin{proof}[\textbf{Proof of Theorem \ref{ode}}]
Having checked localizability for $ \mathbf u_N$, we apply Theorem 2.3 \cite{arous_high-dimensional_2024}. To compute $\mathcal F$, by the above,
\begin{equation*}
    \begin{cases}
    \mathcal F_m = 2\mathbb E_{a_1, a_2}[a_1 f'(a_1 m + a_2 r_\perp) (f(a_1 m + a_2 r_\perp) - f(a_1))],\\
    \mathcal F_{r_\perp^2} = 4r_\perp \mathbb E_{a_1, a_2}[a_2 f'(a_1 m + a_2 r_\perp) (f(a_1 m + a_2 r_\perp) - f(a_1))].
\end{cases}
\end{equation*}

We next turn to calculating the corrector. Recall $V = \mathbb E[\nabla H \otimes \nabla H]$, we have that
\begin{align*}
    V_{ij} = \mathbb E[\partial_i H \partial_j H] = \mathbb  E[\partial_i L_N \partial_j L_N] - \partial_i \Phi\partial_j \Phi,
\end{align*}
where
\begin{align*}
    \nonumber\mathbb E[\partial_i L_N \partial_j L_N] &= 4\mathbb E[a_i a_j f'^2(\langle a, x\rangle) (f(\langle a, x\rangle) - f(a_1))^2] + 4\mathbb E[\epsilon^2 a_i a_j f'^2(\langle a, x\rangle)]\\\nonumber
    &= 4\mathbb E[a_i a_j]\mathbb E[f'^2(\langle a, x\rangle) (f(\langle a, x\rangle) - f(a_1))^2] + 4C_\epsilon \mathbb E[a_i a_j] \mathbb E[f'^2(\langle a, x\rangle)] \\
    & + 4\mathrm{Cov}(a_i a_j, f'^2(\langle a, x\rangle) (f(\langle a, x\rangle) - f(a_1))^2) + 4 \mathrm{Cov}(a_i a_j, f'^2(\langle a, x\rangle)).
\end{align*}
In particular, for $\delta_N = c_\delta/N$, we have $\delta_N \mathcal L_N m = 0$ and 
\begin{align}\label{eq1}
    \delta_N \mathcal L_N r_\perp^2 = \frac{c_\delta}{N}\sum_{i = 2}^N V_{ii} =  \frac{c_\delta}{N}\sum_{i =2}^N \mathbb E[(\partial_i H)^2] =  \frac{c_\delta}{N}\sum_{i =2}^N \mathbb E[(\partial_i L_N)^2] -  \frac{c_\delta}{N}\sum_{i = 2}^N (\partial_i \Phi)^2.
\end{align}
For the first term, one may write
\begin{align}\label{eq2}
    \sum_{i = 2}^N \mathbb E[(\partial_i L_N)^2]&=4(N-1)\left(\mathbb E\left[f'^2(\langle a, x\rangle (f(\langle a, x\rangle) - f(a_1))^2\right]  + C_\epsilon \mathbb E\left[f'^2(\langle a, x\rangle)\right]\right)\\\nonumber&+ 4\sum_{i =2}^N \mathrm{Cov}(a_i^2, f'^2(\langle a, x\rangle (f(\langle a, x\rangle) - f(a_1))^2) +4C_\epsilon \sum_{i = 2}^N\mathrm{Cov}(a_i^2, f'^2(\langle a, x\rangle)).
\end{align}
By the Cauchy-Schwarz inequality, 
\begin{align*}
    \sum_{i =2}^N \mathrm{Cov}(a_i^2, f'^2(\langle a, x\rangle (f(\langle a, x\rangle) - f(a_1))^2) \leq 
\sqrt{2(N-1){\mathrm{Var}(f'^2(\langle a, x\rangle (f(\langle a, x\rangle) - f(a_1))^2)}},
\end{align*}
\begin{align*}
    \sum_{i = 1}^N \mathrm{Cov}(a_i^2, f'(\langle a, x\rangle)) \leq \sqrt{2(N-1) \mathrm{Var}(f^2(\langle a, x\rangle))}.
\end{align*}
This results in the covariance terms being $O_p(\sqrt N)$, which vanishes in terms of the correction when it is multiplied by the step-size $\delta_N = c_\delta/N$. Similarly, the population term, $\sum_{i = 2}^N (\partial_i \Phi)^2 = 4r_\perp^2 (\partial_{r_\perp^2}\Phi)^2 = O_p(1)$, can be seen to vanish in the corrector as $N \to \infty$.

Finally, the corrector is given by
\begin{align*}
    \mathcal G_{r_\perp^2} = 4c_\delta\mathbb E_{a_1, a_2}\left[f'^2(a_1 m + a_2 r_\perp) \left((f(a_1 m + a_2 r_\perp) - f(a_1))^2 + C_\epsilon\right)\right].
\end{align*}
Together, these yield the ODE system for $(m, r_\perp^2)$,
\begin{align}
         \frac{\mathrm{d}m}{\mathrm{d}t} = &-2\mathbb E_{a_1, a_2}[a_1 f'(a_1 m + a_2 r_\perp)(f(a_1 m + a_2 r_\perp) - f(a_1))],\\
         \frac{\mathrm dr_\perp^2}{\mathrm dt} = &-4\mathbb E_{a_1, a_2}[a_2r_\perp f'(a_1 m + a_2 r_\perp)(f(a_1 m + a_2 r_\perp) - f(a_1))] \\\nonumber&+ 4c_\delta\mathbb E_{a_1, a_2}[f'^2(a_1 m+ a_2 r_\perp)((f(a_1 m+ a_2 r_\perp) - f(a_1))^2 + C_\epsilon)].
    \end{align}

Now, let us find the fixed point of $m$ in the above system. Recalling the Hermite polynomials, $(h_k(x))_{k = 0}^\infty$, the activation function can be expressed in terms of Hermite polynomials as follows
\begin{equation*}
    f(x) = \sum_{k}\alpha_k h_k(x) \quad \text{where,} \quad \alpha_k = \langle f, h_k\rangle = \frac{1}{\sqrt{2\pi}}\int_{-\infty}^\infty f(z) h_k(z)e^{-z^2/2}dz.
\end{equation*}
Since $\mathbb E[h_k] = 0$ for all $k \geq  1$, we get $\mathbb E[f(a_1)] = \alpha_0$.  Now, evaluating the ODE at $m = 0$, we obtain the ODE for $m$ given by
\begin{equation*}
    \frac{\mathrm{d}m}{\mathrm{d}t} = -2\mathbb E_{a_1, a_2}[a_1 f'(a_2 r_\perp)f(a_2 r_\perp)] + 2\mathbb E_{a_1, a_2}[a_1 f(a_1)f'(a_2 r_\perp)].
\end{equation*}
Then, using Stein's lemma, one may write
\begin{equation}
     \frac{\mathrm{d}m}{\mathrm{d}t} = 2\mathbb E_{a_1}[f'(a_1)]\mathbb E_{a_2}[f'(a_2 r_\perp)].
\end{equation}

Moreover, $f'(x) = \sum_k \beta_k h_k$ where $\beta_k = (k+1)\alpha_{k+1}$. Hence, $\mathbb E[f'(a_1)] = \alpha_1$ and
\[\frac{\mathrm dm}{\mathrm dt} = 2\alpha_1 \mathbb E_{a_2}[f'(a_2 r_\perp)] = 0.\]
\end{proof}
\begin{proof}[\textbf{Proof of Corollary \ref{cor1}}]
Since $m = 0$ is the initial point of the dynamic, one may write
\begin{align*}
    \frac{\mathrm dr_\perp^2}{\mathrm dt} = &-4\mathbb E_{a_1, a_2}[a_2r_\perp f'(a_2 r_\perp)(f(a_2 r_\perp) - f(a_1))] \\\nonumber&+ 4c_\delta\mathbb E_{a_1, a_2}[f'^2(a_2 r_\perp)((f(a_2 r_\perp) - f(a_1))^2 + C_\epsilon)].
\end{align*}
Using Stein's lemma, one may write
\begin{align}
        \nonumber\frac{\mathrm dr_\perp^2}{\mathrm dt} &= 4\mathbb E_{a_2}[f'^2(a_2 r_\perp)](c_\delta C_\epsilon + c_\delta\|f\|_{L^2}^2 - r_\perp^2) \\
&+4c_\delta\mathbb E_{a_2}[f'^2(a_2 r_\perp)f^2(a_2 r_\perp)]- 4r_\perp^2 \mathbb E_{a_2}[f''(a_2 r_\perp) f(a_2 r_\perp)]\\
& \nonumber+ 4\mathbb E_{a_1}[f(a_1)]\left(r_\perp^2 \mathbb E_{a_2}[f''(a_2 r_\perp)] - 2c_\delta\mathbb E_{a_2}[f'^2(a_2 r_\perp) f(a_2 r_\perp)]\right).
    \end{align}
\end{proof}
\subsection{Proofs of Theorem \ref{sde} and Corollary \ref{cor2}}
\begin{lemma}
    In a single-index model, the distribution of the loss $L_N(x, (a, y))$ depends only on $\tilde{\mathbf{u}}_N = (\sqrt{N}m, r_\perp^2)$. Also, $ \tilde{\mathbf{u}}_N$ is $\delta_N$-localizable for $E_K$ being the centered balls of radius $K$ in $\mathbb R^2$.
\end{lemma}
\begin{proof}
    By rotational invariance of the Gaussian ensemble, we may take $x^* = v$ where $v$ is the first basis vector of $\mathbb R^N$. We have checked localizability in Lemma \ref{loc}, but the change from the original variables is in the $J_N$ matrix, in which now $\nabla \tilde m = \sqrt{N}\nabla m = \sqrt{N} v$. This does not affect the first two conditions of localizability; for the third condition, notice that for some $C > 0$
    \begin{equation*}
        \mathbb E[\langle \nabla H, \nabla \tilde m\rangle^4] = N^2 \mathbb E[\langle \nabla H, v\rangle^4]= N^2 \mathbb E[(\partial_1 H)^4]\leq N^2 C,
    \end{equation*}
    and the second part of the third condition is unchanged since $\nabla^2 \tilde m = 0$.
\end{proof}
\begin{proof}[\textbf{Proof of Theorem \ref{sde}}]
Most of the bounds assumed in the definition of localizability are used to establish tightness and to ensure that higher-order terms in Taylor expansions vanish in the $N \to \infty$ limit. 

Having checked localizability for $\tilde{\mathbf u}_N$, in a neighborhood of $ m = 0$, by Taylor expansion,
    \begin{equation*}
        f(a_1 m + a_2 r_\perp) = f(a_2 r_\perp) + \frac{\tilde m}{\sqrt{N}}a_1 f'(a_2 r_\perp) + \frac{\tilde m^2}{2N}a_1^2 f''(a_2 r_\perp) + O_p(N^{-3/2}).
    \end{equation*}
    Then the population can be expressed as follows
    \begin{align*}
        \phi(\tilde m, r_\perp^2) &= \mathbb E_{a_1, a_2}[(f(a_2 r_\perp) - f(a_1))^2] + \frac{\tilde m^2}{N}\mathbb E_{a_2}[f''(a_2 r_\perp) f(a_2 r_\perp)] \\&+ \frac{\tilde m^2}{N} \mathbb E_{a_2}[f'^2(a_2 r_\perp)] - \frac{\tilde m^2}{N}\mathbb E_{a_1}[a_1^2 f(a_1)]\mathbb E_{a_2} [f''(a_2 r_\perp)] + O(N^{-3/2}).
    \end{align*}
    Note that, by Stein's lemma $\mathbb E_{a_1}[a_1^2 f(a_1)] = \mathbb E[f(a_1) + f''(a_1)]$.
    
    One may write the derivatives for $\mathbf{\tilde u}_N$ as
    \begin{equation*}
        \nabla \tilde m = \sqrt{N}v, \qquad \nabla r_\perp^2 = 2(x-mv).
    \end{equation*}
    Similar to the standard summary statistics, $\nabla^2 \tilde m = 0$, while $\nabla^2 r_\perp^2 = 2(I - vv^T)$, and $\nabla^l r_\perp^2 = 0$ for all $l \geq 3$. It yields that 
    \begin{equation*}
        \langle \nabla \tilde m, \nabla\tilde m\rangle = N, \qquad \langle \nabla \tilde m, \nabla r_\perp^2\rangle = 0, \qquad \langle \nabla r_\perp^2, \nabla r_\perp^2\rangle = 4r_\perp^2
    \end{equation*}
One may apply Theorem 2.3 \cite{arous_high-dimensional_2024}. To compute $\mathcal F$, by the above
    \begin{equation*}
        \begin{cases}
    \mathcal F_{\tilde m} = 2\tilde m \left(\mathbb E_{a_2}[(f'^2(a_2 r_\perp) + f(a_2 r_\perp) f''(a_2 r_\perp)] -E_{a_1}[f(a_1) + f''(a_1)]\mathbb E_{a_2} [f''(a_2 r_\perp)]\right),\\
    \mathcal F_{r_\perp^2} = 4r_\perp \mathbb E_{a_2}[a_2f'(a_2 r_\perp )(f(a_2 r_\perp) - \alpha_0)] .
\end{cases}
    \end{equation*}
In particular, for $\delta_N = c_\delta/N$, we have $\delta_N \mathcal L_N m = 0$. Thus, the corrector $\mathcal G_{m} = 0$. Moreover, in a neighborhood of $m = 0$, the only term that survives in the limit $N \to \infty$ of $\mathcal G_{r^2_\perp}$ is the zeroth-order terms of $f$ and $f'$, i.e.,

\begin{align*}
    \mathcal G_{r_\perp^2} &= 4c_\delta\mathbb E_{a_2}[f'^2(a_2 r_\perp) f^2(a_2 r_\perp)] + 4c_\delta\mathbb E_{a_2}[f'^2(a_2 r_\perp)]\left(\mathbb E_{a_1}[f^2(a_1)] + C_\epsilon\right) \\
    & - 8c_\delta\mathbb E_{a_1}[f(a_1)]\mathbb E_{a_2}[f'^2(a_2 r_\perp)f(a_2 r_\perp)].
\end{align*}

Finally, we consider the volatility of the stochastic process one gets in the limit. Rescaling $J_NVJ_N^T$ by noticing that the rescaling $J_N \mapsto \tilde J_N$ multiplies the $(1, 1)$-entry of $J_N V J_N^T$ by $N$ and its off-diagonal entries by $\sqrt{N}$, one may obtain the entries as follows. Thus,
\begin{equation*}
    J_N = \begin{pmatrix}
        \nabla  m \\
        \nabla  r^2
    \end{pmatrix} = \begin{pmatrix}
        1 & 0 & \cdots & 0\\
        0 & 2x_2 & \cdots & 2x_N
    \end{pmatrix}, \quad \quad J_NVJ_N^T = \begin{pmatrix}
        V_{11} & 2\sum_{i = 2}^N x_i V_{1i}\\
        2\sum_{i = 2}^N x_i V_{1i} & 4\sum_{i = 2}^N x_i x_j V_{ij}
    \end{pmatrix}.
\end{equation*}
    Again, in a neighborhood of $m = 0$, the only term that survives in the limit $N \to \infty$ of $J_N VJ_N^T$ is the zeroth-order terms of $f$ and $f'$. The (1, 2)-entry of the volatility is given by
    \begin{align*}
        \nonumber\frac{1}{2}(J_NVJ_N^T)_{12} &= \sum_{i = 2}^N x_i V_{1i} = \sum_{i = 2}^N x_i \mathbb E[\partial_1 H \partial_i H] \\&= 4\sum_{i = 2}^N \mathbb E_{a}[a_1 a_i x_i f'^2(\langle a, x\rangle) ((f(\langle a, x\rangle) - f(a_1))^2 + C_\epsilon)] \\\nonumber&= 4\mathbb E_{a_1, a_2}[a_1 a_2 r_\perp f'^2(a_2 r_\perp) ((f(a_2 r_\perp) - f(a_1))^2 + C_\epsilon)] ,
    \end{align*}
    and since the rescaled volatility is $\delta_N (\tilde J_N V\tilde J_N^T)_{1, 2} = \frac{c_\delta}{N} \sqrt{N} (J_NVJ_N^T)_{1, 2}$, after taking limits, this term will vanish. We could also say the same reason for the (2, 1) entry. 

    For the (2, 2)-entry, one may write
    \begin{align*}
        \nonumber\frac{1}{4}(J_NVJ_N^T)_{22} &= \sum_{i,j = 2}^N x_i x_j V_{ij} = \sum_{i, j = 2}^N x_i x_j \mathbb E[\partial_i H \partial_j H]\\
        & = 4 \sum_{i, j = 2}^N \mathbb E_{a}[a_i x_i a_j x_j f'^2(\langle a, x\rangle)((f(\langle a, x\rangle) - f(a_1))^2 + C_\epsilon)] \\\nonumber
        & = 4 \mathbb E_{a_2}[a_2^2 r_\perp^2 f'^2(a_2r_\perp)((f(a_2 r_\perp) - f(a_1))^2 + C_\epsilon)].
    \end{align*}
    Again, the rescaled volatility is $\delta_N (\tilde J_N V\tilde J_N^T)_{2, 2} = \frac{c_\delta}{N} (J_NVJ_N^T)_{2, 2}$ vanishes when $N \to \infty$. It means the only surviving entry of the volatility is the (1, 1)-entry that is given by
    \begin{align*}
        (J_NVJ_N^T)_{11} &= V_{11} =  \mathbb E[(\partial_1 H)^2] = 4 \mathbb E_{a_1, a_2}[a_1^2 f'^2(a_2 r_\perp)((f(a_2 r_\perp) - f(a_1))^2 + C_\epsilon)].
    \end{align*}
    Hence, the volatility is of the form
    \begin{align*}
        \Sigma_{11}& = 4c_\delta\mathbb E_{a_2}[f'^2(a_2 r_\perp)] (\mathbb E_{a_1}[a_1^2 f^2(a_1)] + C_\epsilon)+4c_\delta\mathbb E_{a_2}[f'^2(a_2 r_\perp) f^2(a_2 r_\perp)]\\& - 8c_\delta\mathbb E_{a_1}[a_1^2 f(a_1)] \mathbb E[f'^2(a_2 r_\perp)f(a_2 r_\perp)], \qquad \Sigma_{21} = \Sigma_{12} = \Sigma_{22} = 0.
    \end{align*}
    By integration by parts and Stein's lemma, one may write
    \begin{equation*}
        \mathbb E_{a_1}[a_1^2 f^2(a_1)] = \mathbb E_{a_1}[f^2(a_1) + 2f'^2(a_1) + 2f(a_1) f''(a_1)].
    \end{equation*}
    Therefore, 
    \begin{align*}
    \Sigma_{11}& = 4c_\delta\mathbb E_{a_2}[f'^2(a_2 r_\perp)] \big(\|f\|^2_{L^2} + 2\|f'\|^2_{L^2} + 2\langle f, f''\rangle_{L^2} + C_\epsilon\big)\\&+4c_\delta\mathbb E_{a_2}[f'^2(a_2 r_\perp) f^2(a_2 r_\perp)] - 8c_\delta(\mathbb E_{a_1}[f(a_1) + f''(a_1)] ) \mathbb E[f'^2(a_2 r_\perp)f(a_2 r_\perp)].
\end{align*}
    Together, these yield the SDE system for $(\tilde m, r_\perp^2)$,
    \begin{align*}
    \nonumber\mathrm{d}\tilde m = &-2\tilde m \mathbb E_{a_2}[f'^2(a_2 r_\perp) + (f(a_2 r_\perp) - (\alpha_0 + \alpha_2))f''(a_2 r_\perp)] \mathrm{d}t  \\&+ 2\sqrt{c_\delta}\big(\mathbb E_{a_2}[f'^2(a_2 r_\perp)] \big(\|f\|^2_{L^2} + 2\|f'\|^2_{L^2} + 2\langle f, f''\rangle_{L^2} + C_\epsilon\big)\\&\nonumber\qquad+\mathbb E_{a_2}[f'^2(a_2 r_\perp) f^2(a_2 r_\perp)] - 2(\alpha_0 + \alpha_2) \mathbb E[f'^2(a_2 r_\perp)f(a_2 r_\perp)]\big)^{1/2}\mathrm{d}B_t,\\
    \nonumber\frac{\mathrm dr_\perp^2}{\mathrm dt} &= 4\mathbb E_{a_2}[f'^2(a_2 r_\perp)](c_\delta C_\epsilon + c_\delta\|f\|_{L^2}^2 - r_\perp^2) \\
&+4c_\delta\mathbb E_{a_2}[f'^2(a_2 r_\perp)f^2(a_2 r_\perp)]- 4r_\perp^2 \mathbb E_{a_2}[f''(a_2 r_\perp) f(a_2 r_\perp)]\\
& \nonumber+ 4\alpha_0\left(r_\perp^2 \mathbb E_{a_2}[f''(a_2 r_\perp)] - 2c_\delta\mathbb E_{a_2}[f'^2(a_2 r_\perp) f(a_2 r_\perp)]\right).
    \end{align*}
By (\ref{shorthand}), one may rewrite the dynamics as follows,
\begin{align*}
    \mathrm{d}\tilde m &= -2\tilde m\,\big[\mathsf a(r_\perp) + \mathsf b(r_\perp) - (\alpha_0+\alpha_2)\mathsf s(r_\perp)\big] \mathrm{d}t \\
 &\qquad+ 2\sqrt{c_\delta}\sqrt{\mathsf q(r_\perp) - 2(\alpha_0+\alpha_2)\mathsf t(r_\perp) + \mathsf a(r_\perp)\big(\|f\|_{L^2}^2 + 2\|f'\|_{L^2}^2 + 2\langle f, f''\rangle_{L^2} + C_\epsilon\big)}\mathrm{d}B_t,\\
    \frac{\mathrm dr_\perp^2}{\mathrm dt} &= 4\mathsf a(r_\perp)\big(c_\delta C_\epsilon+c_\delta \|f\|_{L^2}^2-r_\perp^2\big) + 4c_\delta\mathsf q(r_\perp) - 4r_\perp^2\mathsf b(r_\perp) + 4\alpha_0\big(r_\perp^2\mathsf s(r_\perp)-2c_\delta\mathsf t(r_\perp)\big).
    \end{align*}
\end{proof}
% \subsection{Proof of Corollary \ref{cor2}}
\begin{proof}[\textbf{Proof of Corollary \ref{cor2}}]
    Since the information exponent of $f$ is at least three, its Hermite coefficients satisfy $\alpha_1 = \alpha_2 = 0$. Substituting $\alpha_2 = 0$, while keeping $\alpha_0 \neq 0$, and evaluating Theorem \ref{sde} at the fixed point $r_\perp^*$ gives the stated drift $\gamma(r_\perp^*)$. It remains to show $\gamma(r_\perp^*) > 0$. 

    Recall (\ref{sde_r}), $\gamma(r_\perp^*) = \mathsf a(r_\perp^*) + \mathsf b(r_\perp^*) - \alpha_0 \mathsf s(r_\perp^*)$. If $\alpha_2 = 0$, one may write,
    \[\mathsf a(r_\perp^*) (c_\delta C_\epsilon + c_\delta \|f\|^2_{L^2} - r_\perp^{*2}) + \mathsf q(r_\perp^*) - r_\perp^{*2} \mathsf b(r_\perp^*) + \alpha_0 (r_\perp^{*2} \mathsf s(r_\perp^*) - 2c_\delta\mathsf t(r_\perp^*)) = 0.\]
    Solving for $r_\perp^{*2} \mathsf b(r_\perp^*)$ and substituting into $\gamma(r_\perp^*)$, we have
    \[\gamma(r_\perp^*) = \frac{c_\delta}{r_\perp^{*2}}(\mathsf a(r_\perp^*)(C_\epsilon + \|f\|^2_{L^2}) + \mathsf q(r_\perp^*) - 2\alpha_0 \mathsf t(r_\perp^*)).\]
    Moreover, 
    \[\mathsf q(r_\perp^*) - 2\alpha_0 \mathsf t(r_\perp^*) = \mathbb E_{a_2} [f'^2(a_2 r_\perp^*)(f(a_2 r_\perp^*) - \alpha_0)^2] - \alpha_0^2 \mathsf a(r_\perp^*),\]
    and using $\|f\|^2_{L^2} - \alpha_0^2 = \mathbb E_{a_1}[(f(a_1) - \alpha_0)^2] \geq 0$, we have
    \begin{equation}\label{theta_final}
        \gamma(r_\perp^*) = \frac{1}{r_\perp^{*2}}\big(\mathbb E_{a_2} [f'^2(a_2 r_\perp^*)](C_\epsilon + \mathbb E_{a_1}[(f(a_1) - \alpha_0)^2]) + \mathbb E_{a_2}[f'^2(a_2 r_\perp^*)(f(a_2 r_\perp^*) - \alpha_0)^2]\big).
    \end{equation}
    On the RHS, the first term is a sum of the noise variance $C_\epsilon$ and $\mathrm{Var}(f(a_1))$, multiplied by $\mathbb E_{a_2}[f'^2(a_2 r_\perp^*)] \geq 0$, and the second term is positive, because it is an expectation of a product of squares, meaning that $\gamma(r_\perp^*) > 0$. Hence, the rescaled correlation $\tilde m$ is mean-reverting to zero near the fixed point $r_\perp^*$.
\end{proof}
\section*{Acknowledgment}
I would like to thank Aukosh Jagannath, who supervised my master's thesis at the University of Waterloo, for his discussions and resources; his guidance influenced both that thesis and this paper. This work was supported by the Natural Sciences and Engineering Research Council of Canada (NSERC). Cette recherche a été enterprise
grâce, en partie, au soutien financier du Conseil de recherches en sciences naturelles et en génie
du Canada (CRSNG), [RGPIN-2020-04597].
\bibliography{references}

@inproceedings{anastasiou_normal_2019,
	title = {Normal {Approximation} for {Stochastic} {Gradient} {Descent} via {Non}-{Asymptotic} {Rates} of {Martingale} {CLT}},
	url = {https://proceedings.mlr.press/v99/anastasiou19a.html},
	language = {en},
	urldate = {2026-04-28},
	booktitle = {Proceedings of the {Thirty}-{Second} {Conference} on {Learning} {Theory}},
	publisher = {PMLR},
	author = {Anastasiou, Andreas and Balasubramanian, Krishnakumar and Erdogdu, Murat A.},
	month = jun,
	year = {2019},
	note = {ISSN: 2640-3498},
	pages = {115--137},
}

@misc{araujo_mean-field_2019,
	title = {A mean-field limit for certain deep neural networks},
	url = {http://arxiv.org/abs/1906.00193},
	doi = {10.48550/arXiv.1906.00193},
	urldate = {2026-04-28},
	publisher = {arXiv},
	author = {Araújo, Dyego and Oliveira, Roberto I. and Yukimura, Daniel},
	month = jun,
	year = {2019},
	note = {arXiv:1906.00193 [math]},
}

@article{arous_online_2021,
	title = {Online stochastic gradient descent on non-convex losses from high-dimensional inference},
	volume = {22},
	issn = {1533-7928},
	url = {http://jmlr.org/papers/v22/20-1288.html},
	number = {106},
	urldate = {2026-04-28},
	journal = {Journal of Machine Learning Research},
	author = {Arous, Gerard Ben and Gheissari, Reza and Jagannath, Aukosh},
	year = {2021},
	pages = {1--51},
}

@article{arous_high-dimensional_2024,
	title = {High-dimensional limit theorems for {SGD}: {Effective} dynamics and critical scaling},
	volume = {77},
	issn = {1097-0312},
	shorttitle = {High-dimensional limit theorems for {SGD}},
	url = {https://onlinelibrary.wiley.com/doi/abs/10.1002/cpa.22169},
	doi = {10.1002/cpa.22169},
	language = {en},
	number = {3},
	urldate = {2026-04-28},
	journal = {Communications on Pure and Applied Mathematics},
	author = {Arous, Gérard Ben and Gheissari, Reza and Jagannath, Aukosh},
	year = {2024},
	note = {\_eprint: https://onlinelibrary.wiley.com/doi/pdf/10.1002/cpa.22169},
	pages = {2030--2080},
}

@inproceedings{benaim_dynamics_1999,
	address = {Berlin, Heidelberg},
	title = {Dynamics of stochastic approximation algorithms},
	isbn = {978-3-540-48407-3},
	doi = {10.1007/BFb0096509},
	language = {en},
	booktitle = {Séminaire de {Probabilités} {XXXIII}},
	publisher = {Springer},
	author = {Benaïm, Michel},
	editor = {Azéma, Jacques and Émery, Michel and Ledoux, Michel and Yor, Marc},
	year = {1999},
	pages = {1--68},
}

@book{benveniste_adaptive_1990,
	address = {Berlin, Heidelberg},
	title = {Adaptive {Algorithms} and {Stochastic} {Approximations}},
	copyright = {http://www.springer.com/tdm},
	isbn = {978-3-642-75896-6 978-3-642-75894-2},
	url = {http://link.springer.com/10.1007/978-3-642-75894-2},
	urldate = {2026-04-28},
	publisher = {Springer},
	author = {Benveniste, Albert and Métivier, Michel and Priouret, Pierre},
	year = {1990},
	doi = {10.1007/978-3-642-75894-2},
}

@article{biehl_learning_1995,
	title = {Learning by on-line gradient descent},
	volume = {28},
	issn = {0305-4470},
	url = {https://doi.org/10.1088/0305-4470/28/3/018},
	doi = {10.1088/0305-4470/28/3/018},
	language = {en},
	number = {3},
	urldate = {2026-04-28},
	journal = {Journal of Physics A: Mathematical and General},
	author = {Biehl, M. and Schwarze, H.},
	month = feb,
	year = {1995},
	pages = {643},
}

@incollection{saad_-line_1999,
	address = {Cambridge},
	series = {Publications of the {Newton} {Institute}},
	title = {On-line {Learning} and {Stochastic} {Approximations}},
	isbn = {978-0-521-11791-3},
	url = {https://www.cambridge.org/core/books/online-learning-in-neural-networks/online-learning-and-stochastic-approximations/58E32E8639D6341349444006CF3D689A},
	urldate = {2026-04-28},
	booktitle = {On-{Line} {Learning} in {Neural} {Networks}},
	publisher = {Cambridge University Press},
	author = {Bottou, Léon},
	editor = {Saad, David},
	year = {1999},
	doi = {10.1017/CBO9780511569920.003},
	pages = {9--42},
}

@misc{celentano_high-dimensional_2026,
	title = {The high-dimensional asymptotics of first order methods with random data},
	url = {http://arxiv.org/abs/2112.07572},
	doi = {10.48550/arXiv.2112.07572},
	urldate = {2026-04-28},
	publisher = {arXiv},
	author = {Celentano, Michael and Cheng, Chen and Montanari, Andrea},
	month = apr,
	year = {2026},
	note = {arXiv:2112.07572 [math]},
}

@inproceedings{chizat_global_2018,
	title = {On the {Global} {Convergence} of {Gradient} {Descent} for {Over}-parameterized {Models} using {Optimal} {Transport}},
	volume = {31},
	url = {https://proceedings.neurips.cc/paper_files/paper/2018/hash/a1afc58c6ca9540d057299ec3016d726-Abstract.html},
	urldate = {2026-04-28},
	booktitle = {Advances in {Neural} {Information} {Processing} {Systems}},
	publisher = {Curran Associates, Inc.},
	author = {Chizat, Lénaïc and Bach, Francis},
	year = {2018},
}

@article{collins-woodfin_hitting_2024,
	title = {Hitting the {High}-dimensional notes: an {ODE} for {SGD} learning dynamics on {GLMs} and multi-index models},
	volume = {13},
	copyright = {https://creativecommons.org/licenses/by/4.0/},
	issn = {2049-8772},
	shorttitle = {Hitting the {High}-dimensional notes},
	url = {https://academic.oup.com/imaiai/article/doi/10.1093/imaiai/iaae028/7826065},
	doi = {10.1093/imaiai/iaae028},
	language = {en},
	number = {4},
	urldate = {2026-04-28},
	journal = {Information and Inference: A Journal of the IMA},
	author = {Collins-Woodfin, Elizabeth and Paquette, Courtney and Paquette, Elliot and Seroussi, Inbar},
	month = sep,
	year = {2024},
	pages = {iaae028},
}

@article{dupuis_stochastic_1989,
	title = {Stochastic {Approximation} and {Large} {Deviations}: {Upper} {Bounds} and w.p.1 {Convergence}},
	volume = {27},
	issn = {0363-0129},
	shorttitle = {Stochastic {Approximation} and {Large} {Deviations}},
	url = {https://epubs.siam.org/doi/10.1137/0327059},
	doi = {10.1137/0327059},
	number = {5},
	urldate = {2026-04-28},
	journal = {SIAM Journal on Control and Optimization},
	author = {Dupuis, Paul and Kushner, Harold J.},
	month = sep,
	year = {1989},
	note = {Publisher: Society for Industrial and Applied Mathematics},
	pages = {1108--1135},
}

@inproceedings{goldt_dynamics_2019,
	title = {Dynamics of stochastic gradient descent for two-layer neural networks in the teacher-student setup},
	volume = {32},
	url = {https://proceedings.neurips.cc/paper_files/paper/2019/hash/cab070d53bd0d200746fb852a922064a-Abstract.html},
	urldate = {2026-04-28},
	booktitle = {Advances in {Neural} {Information} {Processing} {Systems}},
	publisher = {Curran Associates, Inc.},
	author = {Goldt, Sebastian and Advani, Madhu and Saxe, Andrew and Krzakala, Florent and Zdeborová, Lenka},
	year = {2019},
}

@inproceedings{harvey_tight_2019,
	title = {Tight analyses for non-smooth stochastic gradient descent},
	url = {https://proceedings.mlr.press/v99/harvey19a.html},
	language = {en},
	urldate = {2026-04-28},
	booktitle = {Proceedings of the {Thirty}-{Second} {Conference} on {Learning} {Theory}},
	publisher = {PMLR},
	author = {Harvey, Nicholas J. A. and Liaw, Christopher and Plan, Yaniv and Randhawa, Sikander},
	month = jun,
	year = {2019},
	note = {ISSN: 2640-3498},
	pages = {1579--1613},
}

@inproceedings{kushner_asymptotic_1983,
	title = {Asymptotic behavior of stochastic approximation and large deviations},
	url = {https://ieeexplore.ieee.org/document/4047509},
	doi = {10.1109/CDC.1983.269799},
	urldate = {2026-04-28},
	booktitle = {The 22nd {IEEE} {Conference} on {Decision} and {Control}},
	author = {Kushner, Harold J.},
	month = dec,
	year = {1983},
	pages = {75--81},
}

@inproceedings{li_diffusion_2017,
	title = {Diffusion {Approximations} for {Online} {Principal} {Component} {Estimation} and {Global} {Convergence}},
	volume = {30},
	url = {https://proceedings.neurips.cc/paper_files/paper/2017/hash/13f3cf8c531952d72e5847c4183e6910-Abstract.html},
	urldate = {2026-04-28},
	booktitle = {Advances in {Neural} {Information} {Processing} {Systems}},
	publisher = {Curran Associates, Inc.},
	author = {Li, Chris Junchi and Wang, Mengdi and Liu, Han and Zhang, Tong},
	year = {2017},
}

@article{li_stochastic_2019,
	title = {Stochastic {Modified} {Equations} and {Dynamics} of {Stochastic} {Gradient} {Algorithms} {I}: {Mathematical} {Foundations}},
	volume = {20},
	issn = {1533-7928},
	shorttitle = {Stochastic {Modified} {Equations} and {Dynamics} of {Stochastic} {Gradient} {Algorithms} {I}},
	url = {http://jmlr.org/papers/v20/17-526.html},
	number = {40},
	urldate = {2026-04-28},
	journal = {Journal of Machine Learning Research},
	author = {Li, Qianxiao and Tai, Cheng and E, Weinan},
	year = {2019},
	pages = {1--47},
}

@inproceedings{li_validity_2021,
	title = {On the {Validity} of {Modeling} {SGD} with {Stochastic} {Differential} {Equations} ({SDEs})},
	volume = {34},
	url = {https://proceedings.neurips.cc/paper_files/paper/2021/hash/69f62956429865909921fa916d61c1f8-Abstract.html},
	urldate = {2026-04-28},
	booktitle = {Advances in {Neural} {Information} {Processing} {Systems}},
	publisher = {Curran Associates, Inc.},
	author = {Li, Zhiyuan and Malladi, Sadhika and Arora, Sanjeev},
	year = {2021},
	pages = {12712--12725},
}

@article{ljung_analysis_1977,
	title = {Analysis of recursive stochastic algorithms},
	volume = {22},
	issn = {1558-2523},
	url = {https://ieeexplore.ieee.org/document/1101561/},
	doi = {10.1109/TAC.1977.1101561},
	number = {4},
	urldate = {2026-04-28},
	journal = {IEEE Transactions on Automatic Control},
	author = {Ljung, L.},
	month = aug,
	year = {1977},
	pages = {551--575},
}

@article{mandt_stochastic_2017,
	title = {Stochastic {Gradient} {Descent} as {Approximate} {Bayesian} {Inference}},
	volume = {18},
	issn = {1533-7928},
	url = {http://jmlr.org/papers/v18/17-214.html},
	number = {134},
	urldate = {2026-04-28},
	journal = {Journal of Machine Learning Research},
	author = {Mandt, Stephan and Hoffman, Matthew D. and Blei, David M.},
	year = {2017},
	pages = {1--35},
}

@article{mcleish_functional_1976,
	title = {Functional and random central limit theorems for the {Robbins}-{Munro} process},
	volume = {13},
	issn = {0021-9002, 1475-6072},
	url = {https://www.cambridge.org/core/journals/journal-of-applied-probability/article/functional-and-random-central-limit-theorems-for-the-robbinsmunro-process/5789566B5EF41D86FB869BE8DB44724E},
	doi = {10.2307/3212676},
	language = {en},
	number = {1},
	urldate = {2026-04-28},
	journal = {Journal of Applied Probability},
	author = {McLeish, D. L.},
	month = mar,
	year = {1976},
	pages = {148--154},
}

@article{mei_mean_2018,
	title = {A mean field view of the landscape of two-layer neural networks},
	volume = {115},
	url = {https://www.pnas.org/doi/abs/10.1073/pnas.1806579115},
	doi = {10.1073/pnas.1806579115},
	number = {33},
	urldate = {2026-04-29},
	journal = {Proceedings of the National Academy of Sciences},
	author = {Mei, Song and Montanari, Andrea and Nguyen, Phan-Minh},
	month = aug,
	year = {2018},
	note = {Publisher: Proceedings of the National Academy of Sciences},
	pages = {E7665--E7671},
}

@inproceedings{needell_stochastic_2014,
	title = {Stochastic {Gradient} {Descent}, {Weighted} {Sampling}, and the {Randomized} {Kaczmarz} algorithm},
	volume = {27},
	url = {https://proceedings.neurips.cc/paper_files/paper/2014/hash/b3310bba2be31e673a7ded3386994599-Abstract.html},
	urldate = {2026-04-29},
	booktitle = {Advances in {Neural} {Information} {Processing} {Systems}},
	publisher = {Curran Associates, Inc.},
	author = {Needell, Deanna and Srebro, Nathan and Ward, Rachel},
	year = {2014},
}

@article{robbins_stochastic_1951,
	title = {A {Stochastic} {Approximation} {Method}},
	volume = {22},
	issn = {0003-4851, 2168-8990},
	url = {https://projecteuclid.org/journals/annals-of-mathematical-statistics/volume-22/issue-3/A-Stochastic-Approximation-Method/10.1214/aoms/1177729586.full},
	doi = {10.1214/aoms/1177729586},
	language = {en},
	number = {3},
	urldate = {2026-04-29},
	journal = {The Annals of Mathematical Statistics},
	author = {Robbins, Herbert and Monro, Sutton},
	month = sep,
	year = {1951},
	note = {Publisher: Institute of Mathematical Statistics},
	pages = {400--407},
}

@article{rotskoff_trainability_2022,
	title = {Trainability and {Accuracy} of {Artificial} {Neural} {Networks}: {An} {Interacting} {Particle} {System} {Approach}},
	volume = {75},
	copyright = {© 2022 Courant Institute of Mathematics and Wiley Periodicals LLC.},
	issn = {1097-0312},
	shorttitle = {Trainability and {Accuracy} of {Artificial} {Neural} {Networks}},
	url = {https://onlinelibrary.wiley.com/doi/abs/10.1002/cpa.22074},
	doi = {10.1002/cpa.22074},
	language = {en},
	number = {9},
	urldate = {2026-04-29},
	journal = {Communications on Pure and Applied Mathematics},
	author = {Rotskoff, Grant and Vanden-Eijnden, Eric},
	year = {2022},
	note = {\_eprint: https://onlinelibrary.wiley.com/doi/pdf/10.1002/cpa.22074},
	pages = {1889--1935},
}

@article{saad_exact_1995,
	title = {Exact {Solution} for {On}-{Line} {Learning} in {Multilayer} {Neural} {Networks}},
	volume = {74},
	url = {https://link.aps.org/doi/10.1103/PhysRevLett.74.4337},
	doi = {10.1103/PhysRevLett.74.4337},
	number = {21},
	urldate = {2026-04-29},
	journal = {Physical Review Letters},
	author = {Saad, David and Solla, Sara A.},
	month = may,
	year = {1995},
	note = {Publisher: American Physical Society},
	pages = {4337--4340},
}

@article{sirignano_mean_2020,
	title = {Mean field analysis of neural networks: {A} central limit theorem},
	volume = {130},
	issn = {0304-4149},
	shorttitle = {Mean field analysis of neural networks},
	url = {https://www.sciencedirect.com/science/article/pii/S0304414918306197},
	doi = {10.1016/j.spa.2019.06.003},
	number = {3},
	urldate = {2026-04-29},
	journal = {Stochastic Processes and their Applications},
	author = {Sirignano, Justin and Spiliopoulos, Konstantinos},
	month = mar,
	year = {2020},
	pages = {1820--1852},
}

@misc{tan_online_2019,
	title = {Online {Stochastic} {Gradient} {Descent} with {Arbitrary} {Initialization} {Solves} {Non}-smooth, {Non}-convex {Phase} {Retrieval}},
	url = {http://arxiv.org/abs/1910.12837},
	doi = {10.48550/arXiv.1910.12837},
	urldate = {2026-04-29},
	publisher = {arXiv},
	author = {Tan, Yan Shuo and Vershynin, Roman},
	month = oct,
	year = {2019},
	note = {arXiv:1910.12837 [stat]},
}

@article{veiga_phase_2022,
	title = {Phase diagram of {Stochastic} {Gradient} {Descent} in high-dimensional two-layer neural networks},
	volume = {35},
	url = {https://proceedings.neurips.cc/paper_files/paper/2022/hash/939bb847ebfd14c6e4d3b5705e562054-Abstract-Conference.html},
	language = {en},
	urldate = {2026-04-29},
	journal = {Advances in Neural Information Processing Systems},
	author = {Veiga, Rodrigo and Stephan, Ludovic and Loureiro, Bruno and Krzakala, Florent and Zdeborová, Lenka},
	month = dec,
	year = {2022},
	pages = {23244--23255},
}
\end{document}